\documentclass[11pt,a4paper]{article}
\usepackage[hyperref]{emnlp-ijcnlp-2019}
\usepackage{times}
\usepackage{latexsym}

\usepackage{amsmath}
\usepackage{graphicx}
\usepackage{booktabs}
\usepackage{multirow}
\usepackage{enumitem}
\usepackage{colortbl}
\usepackage{xcolor}
\usepackage{url}
\usepackage{amssymb}
\usepackage{amsthm}
\usepackage{mathtools}
\usepackage[ruled,vlined]{algorithm2e}
\usepackage{subcaption}
\usepackage{makecell}

\usepackage{xr}
\newcommand{\com}[1]{}

\newcommand{\reword}[2]{\textcolor{orange}{#2\sout{#1}}}

\renewcommand{\reword}[2]{#2}

\newcommand{\KB}[0]{KnowBert}
\newcommand{\KBS}[0]{KnowBert }
\newcommand{\KWIKI}[0]{KnowBert-Wiki}
\newcommand{\KWORDNET}[0]{KnowBert-WordNet}
\newcommand{\KWANDW}[0]{KnowBert-W+W}
\newcommand{\KWIKIS}[0]{KnowBert-Wiki }
\newcommand{\KWORDNETS}[0]{KnowBert-WordNet }
\newcommand{\KWANDWS}[0]{KnowBert-W+W }

\newcommand{\BBS}[0]{BERT$_\texttt{BASE}$ }
\newcommand{\BB}[0]{BERT$_\texttt{BASE}$}
\newcommand{\BLS}[0]{BERT$_\texttt{LARGE}$ }
\newcommand{\BL}[0]{BERT$_\texttt{LARGE}$}

\aclfinalcopy 

\title{Knowledge Enhanced Contextual Word Representations}

\author{\makecell{Matthew E. Peters$^1$, Mark Neumann$^1$, Robert L. Logan IV$^2$, Roy Schwartz$^{1,3}$,\\
Vidur Joshi$^1$, Sameer Singh$^2$, and Noah A. Smith$^{1,3}$}\\
\\
$^1$Allen Institute for Artificial Intelligence, Seattle, WA, USA\\
$^2$University of California, Irvine, CA, USA\\
$^3$Paul G. Allen School of Computer Science \& Engineering, University of Washington\\
\texttt{
    \hypersetup{urlcolor=black}
    \{\href{mailto:matthewp@allenai.org}{matthewp},\href{mailto:markn@allenai.org}{markn},\href{mailto:roys@allenai.org}{roys},\href{mailto:noah@allenai.org}{noah}\}@allenai.org}\\
    \texttt {
    \hypersetup{urlcolor=black}
    \{\href{mailto:rlogan@uci.edu}{rlogan},\href{mailto:sameer@uci.edu}{sameer}\}@uci.edu
    }
}

\date{}

\begin{document}
\maketitle

\begin{abstract}
Contextual word representations, typically trained on unstructured, unlabeled text, do not contain any explicit grounding to real world entities and are often unable to remember facts about those entities.
We propose a general method to embed multiple knowledge bases (KBs) into large scale models, and thereby enhance their representations with structured, human-curated knowledge.
For each KB, we first use an integrated entity linker to retrieve relevant entity embeddings, then update contextual word representations via a form of word-to-entity attention.
In contrast to previous approaches, the entity linkers and self-supervised language modeling objective are jointly trained end-to-end in a multitask setting that combines a small amount of entity linking supervision with a large amount of raw text.
After integrating WordNet and a subset of Wikipedia into BERT, the knowledge enhanced BERT (\KB) demonstrates improved perplexity, ability to recall facts as measured in a probing task and downstream performance on relationship extraction, entity typing, and word sense disambiguation.
\KB's runtime is comparable to BERT's and it scales to large KBs.
\end{abstract}

\section{Introduction}

Large pretrained models such as ELMo \cite{Peters2018}, GPT \cite{Radford2018}, and BERT \cite{Devlin2018} have significantly improved the state of the art for a wide range of NLP tasks.
These models are trained on large amounts of raw text using self-supervised objectives.
However, they do not contain any explicit grounding to real world entities and as a result have difficulty recovering factual knowledge \cite{Logan2019BaracksWH}.

Knowledge bases (KBs) provide a rich source of high quality, human-curated knowledge that can be used to ground these models.
In addition, they often include complementary information to that found in raw text, and can 
encode factual knowledge that is difficult to learn from selectional preferences either due to infrequent mentions of commonsense knowledge or long range dependencies.

We present a general method to insert multiple KBs into a large pretrained model with a Knowledge Attention and Recontextualization (KAR) mechanism.
The key idea is to explicitly model \textit{entity spans} in the input text and use an entity linker to retrieve relevant entity embeddings from a KB to form knowledge enhanced entity-span representations.
Then, the model recontextualizes the entity-span representations with word-to-entity attention to allow long range interactions between contextual word representations and all entity spans in the context.
The entire KAR is inserted between two layers in the middle of a pretrained model such as BERT.

In contrast to previous approaches that integrate external knowledge into task-specific models with task supervision (e.g., \citealp{Yang2017LeveragingKB,chen2017neural}), our approach learns the entity linkers with self-supervision on unlabeled data. This results in general purpose knowledge enhanced representations that can be applied to a wide range of downstream tasks.

Our approach has several other benefits.
First, it leaves the top layers of the original model unchanged so we may retain the output loss layers and fine-tune on unlabeled corpora while training the KAR. 
This also allows us to simply swap out BERT for \KBS in any downstream application.
Second, by taking advantage of the existing high capacity layers in the original model, the KAR is lightweight, adding minimal additional parameters and runtime.
Finally, it is easy to incorporate additional KBs by simply inserting them at other locations.

\KBS is agnostic to the form of the KB, subject to a small set of requirements (see Sec.~\ref{sec:kg_requirements}).
We experiment with integrating both WordNet \cite{miller1995wordnet} and Wikipedia, thus explicitly adding word sense knowledge and facts about named entities (including those unseen at training time).  However, the method could be extended to commonsense KBs such as ConceptNet \cite{Speer2017ConceptNet5A} or domain specific ones \cite[e.g., UMLS; ][]{Bodenreider2004TheUM}.

We evaluate \KBS with a mix of intrinsic and extrinsic tasks.
Despite being based on the smaller \BBS model, the experiments demonstrate improved masked language model perplexity and ability to recall facts over \BL.
The extrinsic evaluations demonstrate improvements for challenging relationship extraction, entity typing and word sense disambiguation datasets, and often outperform other contemporaneous attempts to incorporate external knowledge into BERT.


\section{Related Work}

\paragraph{Pretrained word representations}
Initial work learning word vectors focused on static word embeddings using multi-task learning objectives ~\cite{collobert2008unified} or
corpus level co-occurence statistics~\cite{mikolov2013efficient,Pennington2014GloveGV}.
Recently the field has shifted toward learning context-sensitive embeddings \cite{Dai2015SemisupervisedSL, Mccann2017, Peters2018, Devlin2018}.
We build upon these by incorporating structured knowledge into these models.

\paragraph{Entity embeddings}
Entity embedding methods produce continuous vector representations from external knowledge sources.
Knowledge graph-based methods optimize the score of observed triples in a knowledge graph.
These methods broadly fall into two categories: translational distance models~\cite{bordes2013translating,wang2014knowledge,lin2015learning,xiao2016one} which use a distance-based scoring function, and linear models~\cite{nickel2011three,yang2014embedding,trouillon2016complex,dettmers2018convolutional} which use a similarity-based scoring function.
We experiment with TuckER~\cite{Balazevic2019TuckERTF} embeddings, a recent linear model which generalizes many of the aforecited models.
Other methods combine entity metadata with the graph \cite{Xie2016RepresentationLO}, use entity contexts \cite{Chen2014AUM,Ganea2017DeepJE}, or a combination of contexts and the KB \cite{Wang2014KnowledgeGA,Gupta2017EntityLV}.
Our approach is agnostic to the details of the entity embedding method and as a result is able to use any of these methods.

\paragraph{Entity-aware language models}
Some previous work has focused on adding KBs to generative language models (LMs) \cite{Ahn2017ANK,yang2016reference,Logan2019BaracksWH} or building entity-centric LMs \cite{ji2017dynamic}.  However, these methods introduce latent variables that require full annotation for training, or marginalization.
In contrast, we adopt a method that allows training with large amounts of unannotated text.

\paragraph{Task-specific KB architectures}
Other work has focused on integrating KBs into neural architectures for specific downstream tasks \cite{Yang2017LeveragingKB,Sun2018OpenDQ,chen2017neural,Bauer2018CommonsenseFG,Mihaylov2018KnowledgeableRE,Wang2019ExplicitUO,yang-etal-2019-enhancing-pre}.
Our approach instead uses KBs to learn more generally transferable representations that can be used to improve a variety of downstream tasks.

\section{\KB}

\KBS incorporates knowledge bases into BERT using the Knowledge Attention and Recontextualization component (KAR).
We start by describing the BERT and KB components. 
We then move to introducing KAR.
Finally, we describe the training procedure, including the multitask training regime for jointly training \KBS and an entity linker.

\subsection{Pretrained BERT}

We describe \KBS as an extension to (and candidate replacement for) BERT, although the method is general and can be applied to any deep pretrained model including left-to-right and right-to-left LMs such as ELMo and GPT.
Formally, BERT accepts as input a sequence of $N$ WordPiece tokens~\cite{sennrich-etal-2016-neural,wu2016google},
$(x_1, \ldots, x_N)$, and computes $L$ layers of $D$-dimensional contextual representations
$\mathbf{H}_i \in \mathbb{R}^{N \times D}$ by successively applying non-linear functions $\mathbf{H}_i = \mathrm{F}_i(\mathbf{H}_{i-1})$. 
The non-linear function is a multi-headed self-attention layer followed by a position-wise multilayer perceptron (MLP) \cite{Vaswani2017AttentionIA}:
\begin{align*}
  &  \mathrm{F}_i(\mathbf{H}_{i-1}) = \\
   & \,\,\, \mathrm{TransformerBlock}(\mathbf{H}_{i-1}) = \\
   & \,\,\, \mathrm{MLP}(\mathrm{MultiHeadAttn}(\mathbf{H}_{i-1}, \mathbf{H}_{i-1}, \mathbf{H}_{i-1})).
\end{align*}
The multi-headed self-attention uses $\mathbf{H}_{i-1}$ as the query, key, and value to allow each vector to attend to every other vector.

BERT is trained to minimize an objective function that combines both next-sentence prediction (NSP) and masked LM log-likelihood (MLM):
\begin{align*}
\mathcal{L}_\text{BERT} = \mathcal{L}_\text{NSP} + \mathcal{L}_\text{MLM}.
\end{align*}
Given two inputs $\mathbf{x}_A$ and $\mathbf{x}_B$, the next-sentence prediction task is binary classification to predict whether $\mathbf{x}_B$ is the next sentence following $\mathbf{x}_A$.
The masked LM objective randomly replaces a percentage of input word pieces with a special \texttt{[MASK]} token and computes the negative log-likelihood of the missing token with a linear layer and softmax over all possible word pieces.

\begin{figure*}[!t]
\centering
\includegraphics[width=0.95\textwidth]{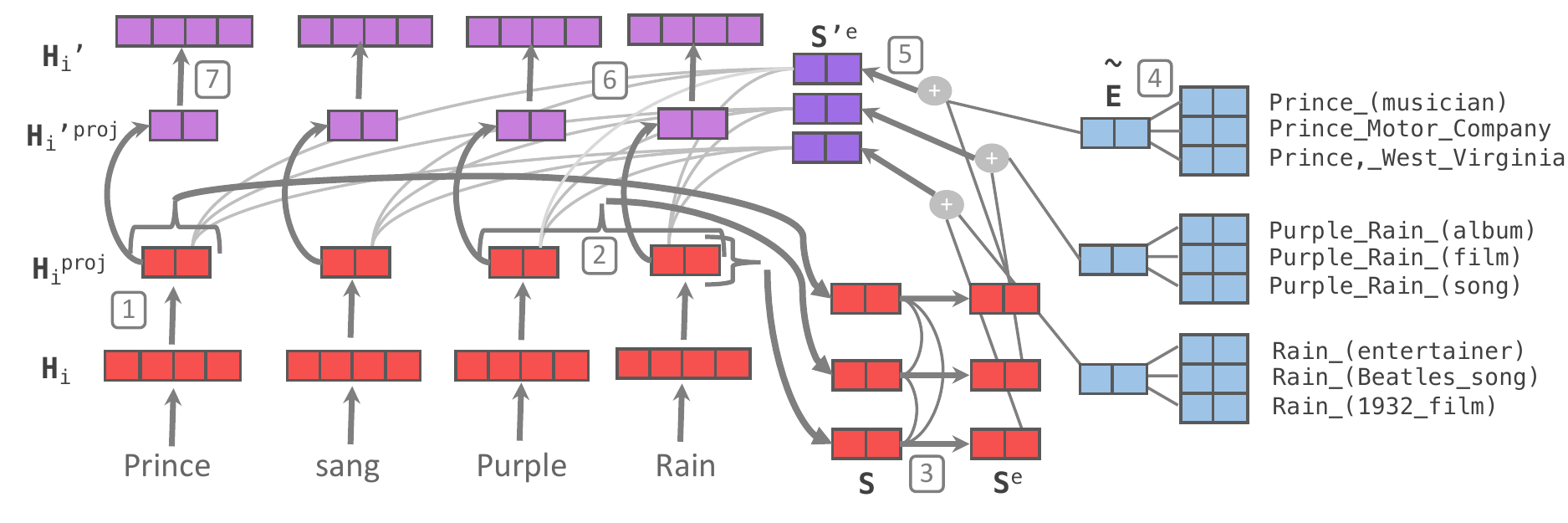}
\caption{The Knowledge Attention and Recontextualization (KAR) component.
BERT word piece representations ($\mathbf{H}_i$) are first projected to $\mathbf{H}^{\text{proj}}_i$ (1), then pooled over candidate mentions spans (2) to compute $\mathbf{S}$, and contextualized into $\mathbf{S}^e$ using mention-span self-attention (3).
An integrated entity linker computes weighted average entity embeddings $\tilde{\mathbf{E}}$ (4), which are used to enhance the span representations with knowledge from the KB (5), computing $\mathbf{S}'^e$.
Finally, the BERT word piece representations are recontextualized with word-to-entity-span attention (6) and projected back to the BERT dimension (7) resulting in $\mathbf{H}'_i$.}
\label{fig:kar}
\end{figure*}

\subsection{Knowledge Bases}
\label{sec:kg_requirements}
The key contribution of this paper is a method to incorporate knowledge bases (KB) into a pretrained BERT model. 
To encompass as wide a selection of prior knowledge as possible, we adopt a broad definition for a KB in the most general sense as fixed collection of $K$ entity nodes, $e_k$, from which it is possible to compute entity embeddings, $\mathbf{e}_k \in \mathbb{R}^E$. 
This includes KBs with a typical \texttt{(subj, rel, obj)} graph structure, KBs that contain only entity metadata without a graph, and those that combine both a graph and entity metadata, as long as there is some method for embedding the entities in a low dimensional vector space.
We also do not make any assumption that the entities are typed.
As we show in Sec.~\ref{sec:exp_setup} this flexibility is beneficial, where we compute entity embeddings from WordNet using both the graph and synset definitions, but link directly to Wikipedia pages without a graph by using embeddings computed from the entity description.

We also assume that the KB is accompanied by an entity candidate selector that takes as input some text and returns a list of $C$ potential entity links, each consisting of the start and end indices of the potential mention span and $M_m$ candidate entities in the KG:
\begin{align*}
\mathcal{C} = \{\langle (\text{start}_m, \text{end}_m), (e_{m,1}, \ldots, e_{m,M_m}) \rangle  \mid  \\ m \in 1 \ldots C, e_k \in 1 \ldots K \}.
\end{align*}
In practice, these are often implemented using precomputed dictionaries \cite[e.g.,\ CrossWikis;][]{spitkovsky-chang-2012-cross}, KB specific rules (e.g., a WordNet lemmatizer), or other heuristics \cite[e.g., string match;][]{Mihaylov2018KnowledgeableRE}.
\citet{Ling2015DesignCF} showed that incorporating candidate priors into entity linkers can be a powerful signal, so we optionally allow for the candidate selector to return an associated prior probability for each entity candidate.
In some cases, it is beneficial to over-generate potential candidates and add a special \texttt{NULL} entity to each candidate list, thereby allowing the linker to discriminate between actual links and false positive candidates.
In this work, the entity candidate selectors are fixed but their output is passed to a learned context dependent entity linker to disambiguate the candidate mentions.

Finally, by restricting the number of candidate entities to a fixed small number (we use 30), \KB's runtime is independent of the size the KB, as it only considers a small subset of all possible entities for any given text.
As the candidate selection is rule-based and fixed, it is fast and in our implementation is performed asynchronously on CPU.
The only overhead for scaling up the size of the KB is the memory footprint to store the entity embeddings.

\subsection{KAR}

The Knowledge Attention and Recontextualization component (KAR) is the heart of \KB.
The KAR accepts as input the contextual representations at a particular layer, $\mathbf{H}_i$, and computes knowledge enhanced representations $\mathbf{H}'_i = \mathrm{KAR}(\mathbf{H}_i, \mathcal{C})$.
This is fed into the next pretrained layer, $\mathbf{H}_{i+1} = \mathrm{TransformerBlock}(\mathbf{H}'_i)$, and the remainder of BERT is run as usual.

In this section, we describe the KAR's key components: mention-span representations, retrieval of relevant entity embeddings using an entity linker, update of mention-span embeddings with retrieved information, and recontextualization of entity-span embeddings with word-to-entity-span attention.
We describe the KAR for a single KB, but extension to multiple KBs at different layers is straightforward.  
See Fig.~\ref{fig:kar} for an overview.

\paragraph{Mention-span representations}  The KAR starts with the KB entity candidate selector that provides a list of candidate mentions which it uses to compute mention-span representations.
$\mathbf{H}_i$ is first projected to the entity dimension ($E$, typically 200 or 300, see Sec.~\ref{sec:exp_setup}) with a linear projection,
\begin{align}
    \mathbf{H}^{\text{proj}}_i = \mathbf{H}_i \mathbf{W_1^{\text{proj}}} + \mathbf{b}_1^\text{proj}.
    \label{eqn:bert_to_kg_proj}
\end{align}
Then, the KAR computes $C$ mention-span representations $\mathbf{s}_m \in \mathbb{R}^E$, one for each candidate mention, by pooling over all word pieces in a mention-span using the self-attentive span pooling from \citet{Lee2017EndtoendNC}.
The mention-spans are stacked into a matrix $\mathbf{S} \in \mathbb{R}^{C \times E}$.

\paragraph{Entity linker}
The entity linker is responsible for performing entity disambiguation for each potential mention from among the available candidates.
It first runs mention-span self-attention to compute
\begin{align}
    \mathbf{S}^e = \mathrm{TransformerBlock}(\mathbf{S}).
    \label{eqn:entity_span_self_attn}
\end{align}
The span self-attention is identical to the typical transformer layer, exception that the self-attention is between mention-span vectors instead of word piece vectors.
This allows \KBS to incorporate global information into each linking decision so that it can take advantage of entity-entity co-occurrence and resolve which of several overlapping candidate mentions should be linked.\footnote{We found a small transformer layer with four attention heads and a 1024 feed-forward hidden dimension was sufficient, significantly smaller than each of the layers in BERT.  Early experiments demonstrated the effectiveness of this layer with improved entity linking performance.}

Following \citet{Kolitsas2018EndtoEndNE}, $\mathbf{S}^e$ is used to score each of the candidate entities while incorporating the candidate entity prior from the KB.
Each candidate span $m$ has an associated mention-span vector $\mathbf{s}_m^e$ (computed via Eq.~\ref{eqn:entity_span_self_attn}), $M_m$ candidate entities with embeddings $\mathbf{e}_{mk}$ (from the KB), and prior probabilities $p_{mk}$.
We compute $M_m$ scores using the prior and dot product between the entity-span vectors and entity embeddings,
\begin{align}
    \psi_{mk} = \mathrm{MLP}(p_{mk}, \mathbf{s}_m^e \cdot \mathbf{e}_{mk}),
    \label{eqn:el_score}
\end{align}
with a two-layer MLP (100 hidden dimensions).

If entity linking (EL) supervision is available, we can compute a loss with the gold entity $e_{mg}$.
The exact form of the loss depends on the KB, and we use both log-likelihood,
\begin{align}
    \mathcal{L}_\mathrm{EL} = -\sum_m \log \left( \frac{\exp(\psi_{mg})}{\sum_k \exp(\psi_{mk})}   \right),
    \label{eqn:el_loss_softmax}
\end{align}
and max-margin,
\begin{align}
 \mathcal{L}_\mathrm{EL}   = & \max(0, \gamma - \psi_{mg})  \,\,\, + \nonumber \\
& \sum_{e_{mk} \neq e_{mg}} \max (0, \gamma + \psi_{mk}),
    \label{eqn:el_loss_margin}
\end{align}
formulations (see Sec.~\ref{sec:exp_setup} for details).

\paragraph{Knowledge enhanced entity-span representations} \KBS next injects the KB entity information into the mention-span representations computed from BERT vectors ($\mathbf{s}^e_m$) to form entity-span representations.
For a given span $m$, we first disregard all candidate entities with score $\psi$ below a fixed threshold, and softmax normalize the remaining scores:
\begin{align*}
    \tilde{\psi}_{mk} = \begin{dcases}
        \frac{\exp(\psi_{mk})}{\sum\limits_{\psi_{mk} \geq \delta} \exp({\psi_{mk})}}, &  \psi_{mk} \geq \delta \\
        0, & \psi_{mk} < \delta.
    \end{dcases}
\end{align*}
Then the weighted entity embedding is
\begin{align*}
\tilde{\mathbf{e}}_m = \sum_k \tilde{\psi}_{mk} \mathbf{e}_{mk}.
\end{align*}
If all entity linking scores are below the threshold $\delta$, we substitute a special \texttt{NULL} embedding for $\tilde{\mathbf{e}}_m$.
Finally, the entity-span representations are updated with the weighted entity embeddings
\begin{align}
    \mathbf{s}'^e_m = \mathbf{s}_m^e + \tilde{\mathbf{e}}_m,
    \label{eqn:span_update}
\end{align}
which are packed into a matrix $\mathbf{S}'^e \in \mathbb{R}^{C \times E}$\reword{as usual}{}.

\paragraph{Recontextualization}
After updating the entity-span representations with the weighted entity vectors, \KBS uses them to recontextualize the word piece representations.
This is accomplished using a modified transformer layer that substitutes the multi-headed self-attention with a multi-headed attention between the projected word piece representations and knowledge enhanced entity-span vectors.
As introduced by \citet{Vaswani2017AttentionIA}, the contextual embeddings $\mathbf{H}_i$ are used for the query, key, and value in multi-headed self-attention.
The word-to-entity-span attention in \KBS substitutes $\mathbf{H}_i^\text{proj}$ for the query, and $\mathbf{S}'^e$ for both the key and value:
\begin{align*}
   \mathbf{H}_i^{'\text{proj}} = 
    \mathrm{MLP}(\mathrm{MultiHeadAttn}(\mathbf{H}^\text{proj}_{i}, \mathbf{S}^{'e}, \mathbf{S}^{'e})).
\end{align*}
This allows each word piece to attend to all entity-spans in the context, so that it can propagate entity information over long contexts.
After the multi-headed word-to-entity-span attention, we run a position-wise MLP analogous to the standard transformer layer.\footnote{As for the multi-headed entity-span self-attention, we found a small transformer layer to be sufficient, with four attention heads and 1024 hidden units in the MLP.}

Finally, ${\mathbf{H}'}_i^{\text{proj}}$ is projected back to the BERT dimension with a linear transformation and a residual connection added:
\begin{align}
    {\mathbf{H}'}_i = {\mathbf{H}'}_i^{\text{proj}} \mathbf{W_2^{\text{proj}}} + \mathbf{b}_2^\text{proj} + \mathbf{H}_i
    \label{eqn:proj2}
\end{align}

\paragraph{Alignment of BERT and entity vectors}
As \KBS does not place any restrictions on the entity embeddings, it is essential to align them with the pretrained BERT contextual representations.
To encourage this alignment we initialize $\mathbf{W_2^{\text{proj}}}$  as the matrix inverse of $\mathbf{W_1^{\text{proj}}}$ (Eq.~\ref{eqn:bert_to_kg_proj}).
The use of dot product similarity (Eq.~\ref{eqn:el_score}) and residual connection (Eq.~\ref{eqn:proj2}) further aligns the entity-span representations with entity embeddings.

\subsection{Training Procedure}
\label{sec:training}
Our training regime incrementally pretrains increasingly larger portions of \KBS before fine-tuning all trainable parameters in a multitask setting with any available EL supervision.  It is similar in spirit to the ``chain-thaw'' approach in \citet{Felbo2017UsingMO}, and is summarized in Alg.~\ref{alg:training}.

\begin{algorithm}[!t]
\SetAlgoLined
\SetArgSty{}
\DontPrintSemicolon
\KwIn{Pretrained BERT and $J$ KBs}
\KwOut{\KB}
\For{$j = 1 \ldots J$}{
 Compute entity embeddings for KB$_j$ \;
 \If{EL supervision available}{
 Freeze all network parameters except those in (Eq.~\ref{eqn:bert_to_kg_proj}--\ref{eqn:el_score}) \;
 Train to convergence using (Eq.~\ref{eqn:el_loss_softmax}) or (Eq.~\ref{eqn:el_loss_margin}) \;
  }
 Initialize $\mathbf{W_2^{\text{proj}}}$ as $(\mathbf{W_1^{\text{proj}}})^{-1}$ \;
 Unfreeze all parameters except entity embeddings \;
 Minimize $\mathcal{L}_\text{\KB} = \mathcal{L}_\text{BERT} +  \sum_{i=1}^{j} \mathcal{L}_{\text{EL}_i}$ }
 \caption{\KBS training method}
 \label{alg:training}
\end{algorithm}

We assume access to a pretrained BERT model and one or more KBs with their entity candidate selectors.
To add the first KB, we begin by pretraining entity embeddings (if not already provided from another source), then freeze them in all subsequent training, including task-specific fine-tuning.
If EL supervision is available, it is used to pretrain the KB specific EL parameters, while freezing the remainder of the network.
Finally, the entire network is fine-tuned to convergence by minimizing
\begin{align*}
    \mathcal{L}_\text{\KB} = \mathcal{L}_\text{BERT} + \mathcal{L}_\text{EL}.
\end{align*}
We apply gradient updates to homogeneous batches randomly sampled from either the unlabeled corpus or EL supervision.

\begin{table*}[th]
\centering
\begin{tabular}{l c c c c c c}
\toprule
\multirow{2}{*}{System} & \multirow{2}{*}{PPL} & Wikidata & \# params.\ & \# params.\ & \# params.\ & Fwd. / Bwd. \\
                  &   & MRR   & masked LM & KAR & entity embed.\ & time \\
 \midrule
 \BBS & 5.5 & 0.09 & 110 & 0 & 0 & 0.25 \\
 \BLS & 4.5 & 0.11 & 336 & 0 & 0 & 0.75 \\
 \KWIKI & 4.3 & 0.26 & 110 & 2.4 & 141 & 0.27 \\
\KWORDNET & 4.1 & 0.22 & 110 & 4.9 & 265 & 0.31 \\
\KWANDW & \textbf{3.5} & \textbf{0.31} & 110 & 7.3 & 406 & 0.33 \\
\bottomrule
\end{tabular}%
\caption{Comparison of masked LM perplexity, Wikidata probing MRR, and number of parameters (in millions) in the masked LM (word piece embeddings, transformer layers, and output layers), KAR, and entity embeddings for BERT and \KB.  The table also includes the total time to run one forward and backward pass (in seconds) on a TITAN Xp GPU (12 GB RAM) for a batch of 32 sentence pairs with total length 80 word pieces.  Due to memory constraints, the \BLS batch is accumulated over two smaller batches.
}
\label{tab:perplexity}
\end{table*}

To add a second KB, we repeat the process, inserting it in any layer above the first one.
When adding a KB, the BERT layers above it will experience large gradients early in training,
as they are subject to the randomly initialized parameters associated with the new KB.
They are thus expected to move further from their pretrained values before convergence compared to parameters below the KB.
By adding KBs from bottom to top, we minimize disruption of the network and decrease the likelihood that training will fail.
See Sec.~\ref{sec:exp_setup} for details of where each KB was added.

The entity embeddings and selected candidates contain lexical information (especially in the case of WordNet), that will make the masked LM predictions significantly easier.
To prevent leaking into the masked word pieces, we adopt the BERT strategy and replace all entity candidates from the selectors with a special \texttt{[MASK]} entity if the candidate mention span overlaps with a masked word piece.\footnote{Following BERT, for 80\% of masked word pieces all candidates are replaced with \texttt{[MASK]}, 10\% are replaced with random candidates and 10\% left unmasked.}
This prevents \KBS from relying on the selected candidates to predict masked word pieces.

\section{Experiments} 

\subsection{Experimental Setup}
\label{sec:exp_setup}
We used the English uncased \BBS model \cite{Devlin2018} to train three versions of \KB: \KWIKI, \KWORDNET, and \KWANDWS that includes both Wikipedia and WordNet.

\paragraph{\KWIKI} The entity linker in \KWIKIS borrows both the entity candidate selectors and embeddings from \citet{Ganea2017DeepJE}.
The candidate selectors and priors are a combination of CrossWikis, a large, precomputed dictionary that combines statistics from Wikipedia and a web corpus \cite{spitkovsky-chang-2012-cross}, and the YAGO dictionary \cite{Hoffart2011RobustDO}.
The entity embeddings use a skip-gram like objective \cite{word2vec} to learn 300-dimensional embeddings of Wikipedia page titles directly from Wikipedia descriptions without using any explicit graph structure between nodes.
As such, nodes in the KB are Wikipedia page titles, e.g., \texttt{Prince\_(musician)}.
\citet{Ganea2017DeepJE} provide pretrained embeddings for a subset of approximately 470K entities.
Early experiments with embeddings derived from Wikidata relations\footnote{\url{https://github.com/facebookresearch/PyTorch-BigGraph}} did not improve results.

We used the AIDA-CoNLL dataset \cite{Hoffart2011RobustDO} for supervision, adopting the standard splits.  This dataset exhaustively annotates entity links for named entities of person, organization and location types, as well as a miscellaneous type.
It does not annotate links to common nouns or other Wikipedia pages.
At both train and test time, we consider all selected candidate spans and the top 30 entities, to which we add the special \texttt{NULL} entity to allow \KBS to discriminate between actual links and false positive links from the selector.
As such, \KBS models both entity mention detection and disambiguation in an end-to-end manner.  Eq.~\ref{eqn:el_loss_margin} was used as the objective.

\paragraph{\KWORDNET} Our WordNet KB combines synset metadata, lemma metadata and the relational graph.
To construct the graph, we first extracted all synsets, lemmas, and their relationships from WordNet 3.0 using the \texttt{nltk} interface.
After disregarding certain symmetric relationships (e.g., we kept the \texttt{hypernym} relationship, but removed the inverse \texttt{hyponym} relationship) we were left with 28 synset-synset and lemma-lemma relationships.
From these, we constructed a graph where each node is either a synset or lemma, and introduced the special \texttt{lemma\_in\_synset} relationship to link synsets and lemmas.
The candidate selector uses a rule-based lemmatizer without part-of-speech (POS) information.\footnote{\url{https://spacy.io/}}

\begin{table}
\centering
\begin{tabular}[t]{l c}
\toprule
System & F$_1$ \\
 \midrule
 WN-first sense baseline & 65.2 \\
 ELMo & 69.2 \\
 \BBS & 73.1 \\
 \BLS & 73.9 \\
 \KWORDNET & 74.9 \\
\KWANDW & \textbf{75.1} \\
\bottomrule
\end{tabular}
\caption{Fine-grained WSD F$_1$.
}
\label{tab:wsd}
\end{table}

Our embeddings combine both the graph and synset glosses (definitions), as early experiments indicated improved perplexity when using both vs.\ just graph-based embeddings.
We used TuckER \cite{Balazevic2019TuckERTF} to compute 200-dimensional vectors for each synset and lemma using the relationship graph.
Then, we extracted the gloss for each synset and used an off-the-shelf state-of-the-art sentence embedding method \cite{Subramanian2018} to produce 2048-dimensional vectors.
These are concatenated to the TuckER embeddings.
To reduce the dimensionality for use in \KB, the frozen 2248-dimensional embeddings are projected to 200-dimensions with a learned linear transformation.

For supervision, we combined the SemCor word sense disambiguation (WSD) dataset \cite{Miller1994UsingAS} with all lemma example usages from WordNet\footnote{To provide a fair evaluation on the WiC dataset which is partially based on the same source, we excluded all WiC train, development and test instances.} and link directly to synsets.
The loss function is Eq.~\ref{eqn:el_loss_softmax}.
At train time, we did not provide gold lemmas or POS tags, so \KBS must learn to implicitly model coarse grained POS tags to disambiguate each word.
At test time when evaluating we restricted \reword{the available }{}candidate entities to just those matching the gold lemma and POS tag, consistent with the standard WSD evaluation.

\paragraph{Training details}
To control for the unlabeled corpus, we concatenated Wikipedia and the Books Corpus \cite{Zhu2015AligningBA} and followed the data preparation process in BERT with the exception of heavily biasing our dataset to shorter sequences of 128 word pieces for efficiency.
Both \KWIKIS and \KWORDNETS insert the KB between layers 10 and 11 of the 12-layer \BBS model.  \KWANDWS adds the Wikipedia KB between layers 10 and 11, with WordNet between layers 11 and 12.
Earlier experiments with \KWORDNETS in a lower layer had worse perplexity.
We generally followed the fine-tuning procedure in \citet{Devlin2018}; see supplemental materials for details.

\begin{table}
\centering
\begin{tabular}{l c c}
\toprule
System & AIDA-A & AIDA-B \\
 \midrule
 \citet{Daiber2013ImprovingEA} & 49.9 & 52.0 \\
 \citet{Hoffart2011RobustDO} & 68.8 & 71.9 \\
 \citet{Kolitsas2018EndtoEndNE} & \textbf{86.6} & \textbf{82.6} \\
 \KWIKI & 80.2 & 74.4\\
 \KWANDW & 82.1 & 73.7 \\
\bottomrule
\end{tabular}
\caption{End-to-end entity linking strong match, micro averaged F$_1$.}
\label{tab:el}

\end{table}

\subsection{Intrinsic Evaluation}
\reword{
This section presents intrinsic evaluations of \KB: perplexity, speed, ability to recall facts, and results on the integrated entity linkers.
}
{}
\paragraph{Perplexity}
Table \ref{tab:perplexity} compares masked LM perplexity for \KBS with \BBS and \BL.
To rule out minor differences due to our data preparation, the BERT models are fine-tuned on our training data before being evaluated.
Overall, \KBS improves the masked LM perplexity, with all \KBS models 
\reword{having perplexity lower than}{outperforming}
\BL, despite being derived from \BB.

\paragraph{Factual recall} To test \KB's ability to recall facts from the KBs, we extracted 90K tuples from Wikidata~\cite{Vrandecic:2014:WFC:2661061.2629489} for 17 different relationships such as \texttt{companyFoundedBy}.
Each tuple was written into natural language such as ``Adidas was founded by Adolf Dassler'' and used to construct two test instances, one that masks out the subject and one that masks the object.
Then, we evaluated whether a model could recover the masked entity by computing the mean reciprocal rank (MRR) of the masked word pieces.
Table \ref{tab:perplexity} displays a summary of the results (see supplementary material for results across all relationship types).
Overall, \KWIKIS is significantly better at recalling facts than both \BBS and \BL, with \KWANDWS better still.

\paragraph{Speed} \KBS is almost as fast as \BBS (8\% slower for \KWIKI, 32\% for \KWANDW) despite adding a 
\reword{very}{}
large number of (frozen) parameters in the entity embeddings (Table \ref{tab:perplexity}).  \KBS is much faster than \BL.
By taking advantage of the already high capacity model, the number of trainable parameters added by \KBS is a 
\reword{small}{} 
fraction of the total parameters in BERT.
The faster speed is partially due to the entity parameter efficiency in \KBS as only as small fraction of parameters in the entity embeddings are used for any given input due to the sparse linker.
Our candidate generators consider the top 30 candidates and produce approximately $O($number tokens$)$ candidate spans.  For a typical 25 token sentence, approximately 2M entity embedding parameters are actually used.  In contrast, \BLS uses the majority of its 336M parameters for each input.

\begin{table}
\centering
\resizebox{\columnwidth}{!}{
\begin{tabular}{l c c c c}
\toprule
System & LM & P & R & F$_1$ \\
 \midrule
 \citet{Zhang2018GraphCO} & --- & 69.9 & 63.3 & 66.4 \\
 \citet{Alt2018ImprovingRE} & GPT & 70.1 & 65.0 & 67.4 \\
 \citet{ShiSimpleBERTRELSRL2019}  & \BBS & 73.3 & 63.1 & 67.8 \\
  \citet{Zhang2019ERNIEEL}  & \BBS & 70.0 & 66.1 & 68.0 \\
 \citet{Soares2019MatchingTB} &\BLS & --- & --- & 70.1 \\ 
 \citet{Soares2019MatchingTB} & \BL$\dagger$ & --- & --- & \textbf{71.5} \\
 \KWANDW & \BBS &  71.6 & 71.4 & \textbf{71.5} \\
\bottomrule
\end{tabular}}
\caption{Single model test set results on the TACRED relationship extraction dataset.  $\dagger$ with MTB pretraining. 
}
\label{tab:tacred}
\end{table}

\paragraph{Integrated EL} It is also possible to evaluate the performance of the integrated entity linkers inside \KBS using diagnostic probes without any further fine-tuning.  As these were trained in a multitask setting primarily with raw text, we do not \textit{a priori} expect high performance as they must balance specializing for the entity linking task and learning general purpose representations suitable for language modeling.

Table \ref{tab:wsd} displays fine-grained WSD F$_1$ using the evaluation framework from \citet{Navigli2017WordSD} and the ALL dataset (combing SemEval 2007, 2013, 2015 and Senseval 2 and 3).
By linking to nodes in our WordNet graph and restricting to gold lemmas at test time we can recast the WSD task under our general entity linking framework.
The ELMo and BERT baselines use a nearest neighbor approach trained on the SemCor dataset, similar to the evaluation in \citet{Melamud2016context2vecLG}, which has previously been shown to be competitive with task-specific architectures \cite{Raganato2017NeuralSL}.  As can be seen, \KBS provides competitive performance, and \KWANDWS is able to match the performance of \KWORDNETS despite incorporating both Wikipedia and WordNet.

Table \ref{tab:el} reports end-to-end entity linking performance for the AIDA-A and AIDA-B datasets.  Here, \KB's performance lags behind the current state-of-the-art model from \citet{Kolitsas2018EndtoEndNE}, but still provides strong performance compared to other established systems such as AIDA \cite{Hoffart2011RobustDO} and DBpedia Spotlight \cite{Daiber2013ImprovingEA}.
We believe this is due to the selective annotation in the AIDA data that only annotates named entities.  The CrossWikis-based candidate selector used in \KBS generates candidate mentions for all entities including common nouns from which \KBS may be learning to extract information, at the detriment of specializing to maximize linking performance for AIDA.

\begin{table}
\centering
\begin{tabular}{l c c c c}
\toprule
System & LM & F$_1$ \\
 \midrule
 \citet{wang-etal-2016-relation} & --- & 88.0 \\
 \citet{wang-etal-2019-extracting} & \BBS & 89.0 \\
 \citet{Soares2019MatchingTB} &\BLS & 89.2 \\ 
 \citet{Soares2019MatchingTB} & \BL$\dagger$ & \textbf{89.5} \\
 \KWANDW & \BBS &  89.1 \\
\bottomrule
\end{tabular}
\caption{Test set F$_1$ for SemEval 2010 Task 8 relationship extraction.
$\dagger$ with MTB pretraining. 
}
\label{tab:semeval2010}
\end{table}

\subsection{Downstream Tasks}

\label{sec:downstream}

This section evaluates \KBS on downstream tasks to validate that the addition of knowledge improves performance on tasks expected to benefit from it.
Given the overall superior performance of \KWANDWS on the intrinsic evaluations, we focus on it exclusively for evaluation in this section.  The main results are included in this section; see the supplementary material for full details.

The baselines we compare against are \BB, \BL, the pre-BERT state of the art, and two contemporaneous papers that add similar types of knowledge to BERT.  ERNIE \cite{Zhang2019ERNIEEL} uses TAGME \cite{Ferragina2010TAGMEOA} to link entities to Wikidata, retrieves the associated entity embeddings, and fuses them into \BBS by fine-tuning.
\citet{Soares2019MatchingTB} learns relationship representations by fine-tuning \BLS with large scale ``matching the blanks'' (MTB) pretraining using entity linked text.

\begin{table}
\centering
\begin{tabular}[t]{l c}
\toprule
System & Accuracy \\
 \midrule
 ELMo$\dagger$ & 57.7 \\
 \BB$^\dagger$ & 65.4 \\
 \BL$^\dagger$ & 65.5 \\
 \BL$^{\dagger\dagger}$ & 69.5 \\
 \KWANDW & \textbf{70.9} \\
\bottomrule
\end{tabular}
\caption{Test set results for the WiC dataset (v1.0). \\
$^\dagger$\citet{Pilehvar2019WiCTW} \\
$^{\dagger\dagger}$\citet{wang2019superglue}
}
\label{tab:wic}
\end{table}

\paragraph{Relation extraction} Our first task is relation extraction using the TACRED  \cite{Zhang2017PositionawareAA} and SemEval 2010 Task 8 \cite{Hendrickx2009SemEval2010T8} datasets.
Systems are given a sentence with marked a subject and object, and asked to predict which of several different relations (or no relation) holds.
Following \citet{Soares2019MatchingTB}, our \KBS model uses special entity tokens \texttt{[E1]}, \texttt{[/E1]}, \texttt{[E2]}, \texttt{[/E2]} to mark the location of the subject and object in the input sentence, then concatenates the contextual word representations for \texttt{[E1]} and \texttt{[E2]} to predict the relationship.
For TACRED, we also encode the subject and object types with special tokens and concatenate them to the end of the sentence.

For TACRED (Table \ref{tab:tacred}), \KWANDWS significantly outperforms the comparable \BBS systems including ERNIE by 3.5\%, improves over \BLS by 1.4\%, and is able to match the performance of the relationship specific MTB pretraining in \citet{Soares2019MatchingTB}.
For SemEval 2010 Task 8 (Table \ref{tab:semeval2010}), \KWANDWS F$_1$ falls between the entity aware \BBS model from \citet{wang-etal-2019-extracting}, and the \BLS model from \citet{Soares2019MatchingTB}.

\paragraph{Words in Context (WiC)} WiC \cite{Pilehvar2019WiCTW} is a challenging task that presents systems with two sentences both containing a word with the same lemma and asks them to determine if they are from the same sense or not.  It is designed to test the quality of contextual word representations.  We follow standard practice and concatenate both sentences with a \texttt{[SEP]} token and fine-tune the \texttt{[CLS]} embedding.  As shown in Table \ref{tab:wic}, \KWANDWS sets a new state of the art for this task, improving over \BLS by 1.4\% and reducing the relative gap to 80\% human performance by 13.3\%.

\begin{table}
\centering
\begin{tabular}{l c c c}
\toprule
System  & P & R & F$_1$ \\
 \midrule
 UFET & 68.8 & 53.3 & 60.1 \\
 \BBS & 76.4 & 71.0 & 73.6 \\
 ERNIE & 78.4 & 72.9 & 75.6 \\
 \KWANDW & 78.6 & 73.7 & \textbf{76.1} \\
\bottomrule
\end{tabular}
\caption{Test set results for entity typing using the nine general types from \cite{choi-etal-2018-ultra}.}
\label{tab:entity_typing}
\end{table}

\paragraph{Entity typing} We also evaluated \KWANDWS using the entity typing dataset from \citet{choi-etal-2018-ultra}.  To directly compare to ERNIE, we adopted the evaluation protocol in \citet{Zhang2019ERNIEEL} which considers the nine general entity types.\footnote{Data obtained from \url{https://github.com/thunlp/ERNIE}}
Our model marks the location of a target span with the special \texttt{[E]} and \texttt{[/E]} tokens and uses the representation of the \texttt{[E]} token to predict the type.
As shown in Table \ref{tab:entity_typing}, \KWANDWS shows an improvement of 0.6\%  F$_1$ over ERNIE and 2.5\% over \BB.

\section{Conclusion}

We have presented an efficient and general method to insert prior knowledge into a deep neural model.
Intrinsic evaluations demonstrate that the addition of WordNet and Wikipedia to BERT improves the quality of the masked LM and significantly improves its ability to recall facts.
Downstream evaluations demonstrate improvements for relationship extraction, entity typing and word sense disambiguation datasets.
Future work will involve incorporating a diverse set of domain specific KBs for specialized NLP applications.

\section*{Acknowledgements}
The authors acknowledge helpful feedback from anonymous reviewers and the AllenNLP team.  This research was funded in part by the NSF under awards IIS-1817183 and CNS-1730158.

\bibliography{references}
\bibliographystyle{acl_natbib}

\appendix

\section{Supplemental Material}

\subsection{KnowBert training details}
We generally followed the fine-tuning procedure in \citet{Devlin2018}, using Adam \cite{Kingma2014AdamAM} with weight decay 0.1 and learning rate schedule that linearly increases then decreases.
Similar to ULMFiT \cite{Howard2018}, we found it beneficial to vary the learning rate in different layers when fine tuning with this simple schedule: the randomly initialized newly added layers in the KAR surrounding each KB had the largest learning rate $\alpha$; all BERT layers below had learning rate $0.25 \alpha$; and BERT layers above had $0.5 \alpha$.
\KWIKIS was fine-tuned for a total of 750K gradient updates, \KWORDNETS for 500K updates, and \KWANDWS for 500K updates (after pretraining \KWIKI).
Fine-tuning was done on a single Titan RTX GPU with batch size of 32 using gradient accumulation for long sequences, except for the final 250K steps of \KWANDWS which was trained with batch size of 64 using 4 GPUs.
At the end of training, masked LM perplexity continued to decrease, suggesting that further training would improve the results.

Due to computational expense, we performed very little hyperparameter tuning during the pretraining stage, and none with the full \KWANDWS model.
All tuning was performed with partial training runs by monitoring masked LM perplexity in the early portion of training and terminating the under performing configurations.
For each \KWORDNETS and \KWIKIS we ran experiments with two learning rates (maximum learning rate of 1e-4 and 2e-4) and chose the best value after a few hundred thousand gradient updates (2e-4 for \KWORDNETS and 1e-4 for \KWIKI).

When training \KWORDNETS and \KWIKIS we chose random batches from the unlabeled corpus and annotated entity linking datasets with a ratio of 4:1 (80\% unlabeled, 20\% labeled).
For \KWANDWS we used a 85\%/7.5\%/7.5\% sampling ratio.

\subsection{Task fine-tuning details}

This section details the fine tuning procedures and hyperparameters for each of the end tasks.
All optimization was performed with the Adam optimizer with linear warmup of learning rate over the first 10\% of gradient updates to a maximum value, then linear decay over the remainder of training.   Gradients were clipped if their norm exceeded 1.0, and weight decay on all non-bias parameters was set to 0.01.
Grid search was used for hyperparameter tuning (maximum values bolded below), using five random restarts for each hyperparameter setting for all datasets except TACRED (which used a single seed).
Early stopping was performed on the development set.  Batch size was 32 in all cases.

\paragraph{TACRED}
This dataset provides annotations for 106K sentences with typed subject and object spans and relationship labels across 41 different classes (plus the no-relation label).
The hyperparameter search space was:
\begin{itemize}
    \item learning rate: [\textbf{3e-5}, 5e-5]
    \item number epochs: [1, 2, \textbf{3}, 4, 5]
    \item $\beta_2$: [\textbf{0.98}, 0.999]
\end{itemize}
Maximum development micro F$_1$ is 71.7\%.

\paragraph{SemEval 2010 Task 8}
This dataset provides annotations for 10K sentences with untyped subject and object spans and relationship labels across 18 different classes (plus the no-relation label).
As the task does not define a standard development split, we randomly sampled 500 of the 8000 training examples for development.
The hyperparameter search space was:
\begin{itemize}
    \item learning rate: [\textbf{3e-5}, 5e-5]
    \item number epochs: [1, 2, \textbf{3}, 4, 5, 6, 7, 8, 9, 10]
\end{itemize}
with $\beta_2 = 0.98$.  We used the provided \texttt{semeval2010\_task8\_scorer-v1.2.pl} script to compute F$_1$.  The maximum development F$_1$ averaged across the random restarts was 89.1 $\pm$ 0.77 (maximum value was 90.5 across the seeds).

\begin{table*}[t]
\centering
\begin{tabular}{l c c | c c c}
\toprule
 & \multicolumn{2}{c|}{BERT}  & \multicolumn{2}{c}{KnowBert} \\
 & base & large & Wiki & Wordnet & W+W \\
 \midrule
 companyFoundedBy & 0.08 & 0.08 & 	0.23 & 0.21 & 0.28 \\
movieDirectedBy	& 0.05 & 0.04 & 0.08 & 0.09 & 0.10\\
movieStars &	0.06 &	0.05&	0.09& 0.09&	0.11\\
personCityOfBirth&	0.18&	0.20&	0.33& 0.35 & 0.40\\
personCountryOfBirth&	0.16&	0.17&	0.33& 0.36 &	0.47 \\
personCountryOfDeath&	0.18&	0.18&	0.30& 0.30 &	0.44 \\
personEducatedAt&	0.11&	0.10	&0.37& 0.29 &	0.43 \\
personEmployer&	0.03& 	0.02&	0.20& 0.13 &	0.21 \\
personFather&	0.15&	0.14&	0.42& 0.40 &	0.48 \\
personMemberOfBand&	0.11&	0.11&	0.19 &0.18	&0.23 \\
personMemberOfSportsTeam&	0.01&	0.03 & 0.05	&0.08&	0.13\\
personMother&	0.11&	0.11&	0.28& 0.24 &	0.32\\
personOccupation&	0.14&	0.24&	0.24& 0.24 &	0.30\\
personSpouse&	0.05&	0.06&	0.12& 0.08 &	0.18\\
songPerformedBy&	0.07&	0.05&	0.10& 0.10 &	0.13\\
videoGamePlatform	&0.16&	0.22&	0.27& 0.20 &	0.34\\
writtenTextAuthor&	0.06&	0.05&	0.15& 0.16 &	0.18\\
\midrule
Total &	0.09	& 0.11 &	0.26 & 0.22	& 0.31\\
\bottomrule
\end{tabular}%
\caption{Full results on the Wikidata probing task including all relations.
}
\label{tab:kg_probe_full}
\end{table*}

\paragraph{WiC}
WiC is a binary classification task with 7.5K annotated sentence pairs.
Due to the small size of the dataset, we found it helpful to use model averaging to reduce the variance in the development accuracy across random restarts.
The hyperparameter search space was:
\begin{itemize}
    \item learning rate: [\textbf{1e-5}, 2e-5, 3e-5, 5e-5]
    \item number epochs: [2, 3, 4, \textbf{5}]
    \item $beta_2$: [0.98, \textbf{0.999}]
    \item weight averaging decay: [no averaging, \textbf{0.95}, 0.99]
\end{itemize}
The maximum development accuracy was 72.6.

\paragraph{Entity typing}
As described in Section 4.3, we evaluated on a subset of data corresponding to entities classified by nine different general classes: person, location, object, organization, place, entity, object, time, and event.
$\beta_2$ was set to 0.98.
The hyperparameter search space was:
\begin{itemize}
    \item learning rate: [2e-5, \textbf{3e-5}]
    \item number epochs: [2, 3, 4, 5, 6, 7, 8, 9, \textbf{10}, 11, 12, 13, 14]
    \item $beta_2$: [0.98, \textbf{0.999}]
    \item weight averaging decay: [\textbf{no averaging}, 0.99]
\end{itemize}
The maximum development F$_1$ was 75.5 $\pm$ 0.38 averaged across five seeds, with maximum value of 76.0.

\subsection{Wikidata probing results}

Table \ref{tab:kg_probe_full} shows the results for the Wikidata probing task for all relationships.

\end{document}